# DISTRIBUTED REVISION OF BELIEF COMMITMENT IN MULTI-HYPOTHESES INTERPRETATION *

*Judea Pearl*

Cognitive Systems Laboratory, UCLA Computer Science Department, L.A., CA. 90024

**ABSTRACT:** This paper extends the applications of belief-networks models to include the revision of belief **commitments**, i.e., the categorical instantiation of a subset of hypotheses which constitute the most satisfactory explanation of the evidence at hand. We show that, in singly-connected networks, the most satisfactory explanation can be found in linear time by a message-passing algorithm similar to the one used in belief updating. In multiply-connected networks, the problem may be exponentially hard but, if the network is sparse, topological considerations can be used to render the interpretation task tractable. In general, finding the most probable combination of hypotheses is no more complex than computing the degree of belief for any individual hypothesis.

## 1. Introduction

Human-generated interpretations are normally cast in categorical terms, often involving not just one, but a whole set of propositions which, stated together, offer a satisfactory explanation of the observed data. For example, a physician might state, "This patient apparently suffers from two simultaneous disorders $A$ and $B$ which, due to condition $C$, caused the deterioration of organ $D$." Except for the hedging term "apparently," such a statement conveys a sense of unreserved commitment (of beliefs) to a set of four hypotheses. The individual components in the explanation above are meshed together by mutually enforced cause-effect relationships, forming a cohesive whole; the removal of any one component from the discourse would tarnish the completeness of the entire explanation.

Such a sense of cohesiveness normally suggests that a great amount of refuting evidence would have to be gathered before the current interpretation would undergo a revision. Moreover, once a revision is activated, it will likely change the entire content of the interpretation, not merely its level of plausibility. Another characteristic of coherent explanations is that they do not assign degrees of certainty to any individual hypothesis in the argument; neither do they contain information about alternative, next-to-best combinations of hypotheses.

These behavioral features are somewhat at variance with past work on belief network models of evidential reasoning [Pearl, 1985a]. Thus far, this work has focussed on the task of *belief updating*, i.e., assigning each hypothesis in a network a degree of belief, *BEL* (•), consistent with all observations. The function *BEL* changes smoothly and incrementally with each new item of evidence.

This paper extends the applications of Bayesian analysis and belief networks models to include revision of belief *commitments*, i.e., the categorical acceptance of a subset of hypotheses which, together, constitute the most satisfactory explanation of the evidence at hand. Using probabilistic terminology, that task amounts to finding the most probable instantiation of all hypothesis variables, given the observed data.

In principle, this task seems intractable because enumerating and rating all possible instantiations is computationally prohibitive, and many heuristic techniques have been developed in various fields of application. In pattern recognition the problem became known as the "multimembership problem" [Ben Bassat 1980]; in medical diagnosis it is known as "multiple disorders" [Ben Bassat et al. 1980; Pople 1982; Reggia, Nau & Wang 1983; Cooper 1984; Peng & Reggia 1986] and in circuit diagnosis as "multiple-faults" [Reiter 1985; deKleer & Williams 1986].

This paper departs from previous work by emphasizing a *distributed* computation approach to belief revision. The impact of each new piece of evidence is viewed as a perturbation that propagates through the network via local communication among neighboring concepts, with minimum external supervision. At equilibrium, each variable will be bound to a definite value which, together with all other value assignments, is the best interpretation of the evidence. The main reason for adopting this distributed message-passing paradigm is that it leads to a "transparent" belief revision process in which the intermediate steps are conceptually meaningful, thus establishing confidence in the final result. Additionally, it facilitates the generation of qualitative justifications by tracing the sequence of operations along the activated pathways and then, using their causal or diagnostic semantics, translating them into appropriate verbal expressions.

We show that, in singly-connected networks, the most satisfactory explanation can be found in linear time by a message-passing algorithm similar to the one used in belief updating. In multiply-connected networks, the problem may be exponentially hard but, if the network is sparse, topological considerations can be used to render the interpretation task tractable. In general, assembling the most believable combination of hypotheses is no more complex than computing the degree of belief for any individual hypothesis.

## 2. Review of Belief Updating in Bayesian Belief Networks

Bayesian belief networks are directed acyclic graphs in which the nodes represent propositional variables, the arcs signify the existence of direct causal influences between the linked propositions and the strength of these influences are quantified by conditional probabilities. Thus, if the nodes in the graph represent the ordered variables $X_1, X_2, \cdots, X_n$, then each variable $X_i$ draws arrows from a subset $S_i$ of variables perceived to be "direct causes" of $X_i$, i.e., $S_i$ is the smallest set of $X_i$'s predecessors satisfying $P(x_i \mid S_i) = P(x_i \mid x_1, x_2, \cdots x_{i-1})$. A complete and consistent parametrization of the model can be obtained by specifying, for each $X_i$, an assessment $\hat{P}(x_i \mid S_i)$ of $P(x_i \mid S_i)$. The product of all these local assessments,

$$\hat{P}(x_1, x_2, \cdots x_n) = \Pi_i \, \hat{P}(x_i \mid S_i),$$

constitutes a joint-probability model consistent with the assessed quantities. That is, if we compute the conditional probabilities

* This work was supported in part by the National Science Foundation, Grant DCR 83-13875.

$P(x_i | S_i)$ dictated by $\hat{P}(x_1, x_2, \cdots, x_n)$, the original assessments, $\hat{P}(x_i | S_i)$, are recovered. Thus, for example, the distribution corresponding to the network of Figure 1 can be written by inspection:

$$P(x_1, \cdots, x_6) = P(x_6 | x_5) P(x_5 | x_2, x_3)$$

$$P(x_4 | x_1, x_2) P(x_3 | x_1) P(x_2 | x_1) P(x_1)$$

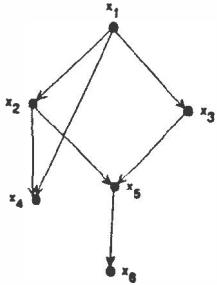

*Figure 1*

A Bayesian network provides a clear graphical representation for the essential independence relationships embedded in the underlying causal model. These independencies can be detected by the following *di-graph separation* criterion: if all paths between $X_i$ and $X_j$ are "blocked" by a subset $S$ of variables, then $X_i$ is independent of $X_j$, given the values of the variables in $S$. A path is "blocked" by $S$ if it contains a member of $S$ between two diverging or two cascaded arrows or, alternatively, if it contains two arrows converging at node $X_k$, and neither $X_k$ nor any of its descendants is in $S$. In particular, each variable $X_i$ is independent of both its grandparents and its non-descendant siblings, given the values of the variables in its parent set $S_i$. In Figure 1, for example, $X_2$ and $X_3$ are independent, given either $\{X_1\}$ or $\{X_1, X_4\}$, because the two paths between $X_2$ and $X_3$ are blocked by either one of these sets. However, $X_2$ and $X_3$ may not be independent given $\{X_1, X_6\}$ because $X_6$, as a descendant of $X_5$, "unblocks" the head-to-head connection at $X_5$, thus opening a pathway between $X_2$ and $X_3$.

Once a Bayesian network is constructed, it can be used as an interpretation engine, namely, newly arriving information will set up a parallel constraint-propagation process which ripples multidirectionally through the networks until, at equilibrium, every variable is assigned a measure of belief consistent with the axioms of probability calculus. Incoming information may be of two types: *specific evidence* and *virtual evidence*. Specific evidence corresponds to direct observations which validate, with certainty, the values of some variables already in the network. Virtual evidence corresponds to judgment based on undisclosed observations which affect the belief of some variables in the network. Such evidence is modeled by dummy nodes representing the undisclosed observations connected by unquantified dummy links to the variables affected by the observations. These links will carry one-way information only -- from the evidence to the variables affected by it.

The objective of updating beliefs coherently by purely local computations can be fully realized if the network is singly-connected, namely, if there is only one undirected path between any pair of nodes. These include causal trees, where each node has a single parent, as well as networks with multi-parent nodes, representing events with several causal factors. We shall first review the propagation scheme in singly-connected networks and then discuss how it can be modified to handle loops.

Let variable names be denoted by capital letters, e.g., $U, V, X, Y, Z$ and their associated values by lower case letters, e.g., $u, v, x, y, z$. All incoming information, both specific and virtual, will be denoted by $e$ to connote *evidence* and will be represented by nodes whose values are precisely known. For the sake of clarity, we will distinguish between the fixed conditional probabilities that label the links, e.g., $P(x | u, v)$, and the dynamic values of the updated node probabilities. The latter will be denoted by $BEL(x)$, which reflects the overall belief accorded to the proposition $X = x$ by all evidence so far received. Thus,

$$BEL(x) \triangleq P(x | e) \qquad (1)$$

where $e$ is the value combination of all instantiated variables.

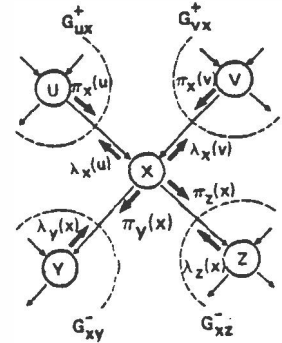

*Figure 2*

Consider a fragment of a singly-connected Bayesian network, as depicted in Figure 2. The link $U \to X$ partitions the graph into two: an upper sub-graph, $G_{ux}^+$, and a lower sub-graph, $G_{ux}^-$, the complement of $G_{ux}^+$. Each of these two sub-graphs may contain a set of evidence, which we shall call respectively $e_{ux}^+$ and $e_{ux}^-$. Likewise, the links $V \to X, X \to Y$ and $X \to Z$ respectively define the sub-graphs $G_{vx}^+, G_{xy}^-$, and $G_{xz}^-$, which may contain the respective evidence sets $e_{vx}^+, e_{xy}^-$ and $e_{xz}^-$.

The belief distribution of each variable $X$ in the network can be computed if three types of parameters are made available:

(1) the current strength of the *causal* support, $\pi$, contributed by each incoming link to $X$:

$$\pi_x(u) = P(u | e_{ux}^+) \qquad (2)$$

(2) the current strength of the *diagnostic* support, $\lambda$, contributed by each outgoing link from $X$:

$$\lambda_y(x) = P(e_{xy}^- | x) \qquad (3)$$

(3) the fixed conditional probability matrix, $P(x | u, v)$, which relates the variable $X$ to its immediate parents.

Using these parameters, local belief updating can be accomplished by the following propagation scheme:

*Step 1:* When node $X$ is activated to update its parameters, it simultaneously inspects the $\pi_x(u)$ and $\pi_x(v)$ communicated by its parents and the messages $\lambda_y(x), \lambda_z(x) \cdots$ communicated by each of its sons. Using this input, it then updates its belief measure as follows:

$$BEL(x) = \alpha \lambda_y(x) \lambda_z(x) \sum_{u,v} P(x | u, v) \pi_x(u) \pi_x(v) \qquad (4)$$

where $\alpha$ is a normalizing constant, rendering $\sum_x BEL(x) = 1$.

*Step 2:* Using the messages received, each node computes new $\lambda$ messages to be sent to its parents. For example, the new message $\lambda_z(u)$ that $X$ sends to its parent $U$ is computed by:

$$\lambda_z(u) = \alpha \sum_v [\pi_x(v) \sum_x \lambda_y(x) \lambda_z(x) P(x|u,v)] \quad (5)$$

*Step 3:* Each node computes new $\pi$ messages to be sent to its children. For example, the new $\pi_y(x)$ message that $X$ sends to its child $Y$ is computed by:

$$\pi_y(x) = \alpha \lambda_z(x) [\sum_{u,v} P(x|u,v) \pi_x(u) \pi_x(v)] \quad (6)$$

The computations described in Eqs. (4), (5) and (6) preserve the probabilistic meaning of the parameters involved. In particular, we have:

$$\lambda_x(u) = P(e_{ux}^-|u) \quad (7)$$

$$\pi_y(x) = P(x|e_{xy}^+) \quad (8)$$

$$BEL(x) = P(x|e) \quad (9)$$

### 3. Belief Commitment in Singly-Connected Networks

Let $W$ stand for the set of all variables considered, explicit as well as virtual, and let $e$ stand for the subset of variables whose values are known precisely at any given time. Our problem is to find an instantiation $w^*$ of all the variables in $W$ which maximizes the conditional probability $P(w|e)$. In other words, $W = w^*$ is the most likely interpretation (MLI) of the evidence at hand:

$$P(w^*|e) = \max_w P(w|e) \quad (10)$$

We consider again the fragment of a singly-connected network in Figure 2 and denote by $W_{xy}^+$ and $W_{xy}^-$ the subset of variables contained in the respective sub-graphs $G_{xy}^+$ and $G_{xy}^-$. Removing any node $X$ would partition the network into the sub-graphs $G_x^+$ and $G_x^-$ containing two sets of variables, $W_x^+$ and $W_x^-$, and (possibly) two sets of evidence, $e_x^+$ and $e_x^-$, respectively.

Using this notation, we can write

$$P(w^*|e) = \max_{x,w_x^+,w_x^-} P(w_x^+, w_x^-, x|e_x^+, e_x^-) \quad (11)$$

The conditional independence of $W_x^+$ and $W_x^-$, given $X$, and the entailments $e_x^+ \subseteq W_x^+$ and $e_x^- \subseteq W_x^-$ yield:

$$P(w^*|e) = \max_{x,w_x^+,w_x^-} \frac{P(w_x^+, w_x^-, x)}{P(e_x^+, e_x^-)}$$

$$= \alpha \max_{x,w_x^+,w_x^-} P(w_x^-|x) P(x|w_x^+) P(w_x^+) \quad (12)$$

where $\alpha = [P(e_x^+, e_x^-)]^{-1}$ is a constant, independent of the uninstantiated variables in $W$. Equation (12) can be rewritten as a maximum, over $x$, of two factors:

$$P(w^*|e) = \alpha \max_x [\max_{w_x^-} P(w_x^-|x)] [\max_{w_x^+} P(x|w_x^+) P(w_x^+)]$$

$$= \alpha \max_x \lambda^*(x) \pi^*(x) \quad (13)$$

where

$$\lambda^*(x) = \max_{w_x^-} P(w_x^-|x) \quad (14)$$

$$\pi^*(x) = \max_{w_x^+} P(x, w_x^+) \quad (15)$$

Thus, if an oracle were to provide us the MLI values of $X$'s descendants $(W_x^-)$, together with the MLI values of all non-descendants of $X$, we would be able to determine the best value of $X$ by computing $\lambda^*(x)$, and $\pi^*(x)$ and then maximize their product, $\lambda^*(x) \pi^*(x)$. This product represents the probability distribution of variable $X$, assuming all other variables are already at their optimal values.

We now express $\lambda^*(x)$ and $\pi^*(x)$ in such a way that they can be computed at node $X$ from similar parameters available at $X$'s neighbors. Writing

$$W_x^- = W_{xy}^- \cup W_{xz}^- \qquad W_x^+ = W_{ux}^+ \cup W_{vx}^+$$
$$W_{u'x}^+ = W_{ux}^+ - U \qquad W_{v'x}^+ = W_{vx}^+ - U,$$

we obtain

$$\lambda^*(x) = \max_{w_{xy}^-} P(w_{xy}^-|x) \max_{w_{xz}^-} P(w_{xz}^-|x) = \lambda_y^*(x) \lambda_z^*(x) \quad (16)$$

and

$$\pi^*(x) = \max_{u,v,w_u^+,w_v^+} [P(x|u,v) P(u,v,w_u^+,w_v^+)]$$

$$= \max_{u,v} [P(x|u,v) \max_{w_{u'x}^+} P(u, w_{u'x}^+) \max_{w_{v'x}^+} P(v, w_{v'x}^+)]$$

$$= \max_{u,v} P(x|u,v) \pi_x^*(u) \pi_x^*(v) \quad (17)$$

where $\lambda_y^*(x)$ (and, correspondingly, $\lambda_z^*(x)$) can be regarded as a message that a child, $Y$, sends to its parent, $X$:

$$\lambda_y^*(x) = \max_{w_{xy}^-} P(w_{xy}^-|x) \quad (18)$$

Similarly,

$$\pi_x^*(u) = \max_{w_{u'x}^+} P(u, w_{u'x}^+) \quad (19)$$

can be regarded as a message that a parent $U$ sends to its child $X$. Note the similarities between $\lambda^*$-$\pi^*$ and $\lambda$-$\pi$ in Eqs. (2) and (3).

Clearly, if these $\lambda^*$ and $\pi^*$ messages are available to $X$, it can compute its best value $x^*$ using Eqs.(13-15). What we must show now is that, upon receiving these messages, it can send back to its neighbors the appropriate $\lambda_x^*(u)$, $\lambda_x^*(v)$, $\pi_y^*(x)$ and $\pi_z^*(x)$ messages, while preserving their probabilistic definitions according to Eqs. (18) and (19).

**Updating $\pi^*$**

Define

$$BEL^*(x) \triangleq P(x, w_x^{+*}, w_x^{-*}|e) \quad (20)$$

203

Using Eqs.(14-19), we can write

$$BEL^*(x) = \alpha \lambda^*(x) \pi^*(x)$$

$$= \alpha \lambda_y^*(x) \lambda_z^*(x) \max_{u,v} P(x|u,v) \pi_x^*(u) \pi_x^*(v) \quad (21)$$

Comparing this expression to the definition of $\pi_y^*(x)$:

$$\pi_y^*(x) = \max_{w_{x'y}^+} P(x, w_{x'y}^+) = \max_{w_x^+, w_{xz}^-} P(x, w_x^+, w_{xz}^-)$$

$$= \alpha \lambda_z^*(x) \max_{u,v} P(x|u,v) \pi_x^*(u) \pi_x^*(v) \quad (22)$$

we see that $\pi_y^*(x)$ can be obtained from $BEL^*(x)$ by setting $\lambda_y^*(x) = 1$ for all $x$. Thus,

$$\pi_y^*(x) = BEL^*(x) \Big|_{\lambda_y(x)=1} = \alpha \frac{BEL^*(x)}{\lambda_y^*(x)} \quad (23)$$

The division by $\lambda_y^*(x)$ in Eq.(23) amounts to discounting the contribution of all variables in $G_{xy}^-$. Note that $\pi_y^*(x)$, unlike $\pi_y(x)$, need not sum to unity over $x$.

Updating $\lambda^*$

Starting with the definition

$$\lambda_x^*(u) = \max_{w_{ux}^-} P(w_{ux}^-|u) \quad (24)$$

we partition $W_{ux}^-$ into its constituents

$$W_{ux}^- = X \cup W_{xy}^- \cup W_{xz}^- \cup W_{v'x}^+ \cup V$$

and obtain

$$\lambda_x^*(u) = \max_{x, w_{xy}^-, w_{xz}^-, w_{v'x}^+, v} P(x, w_{xy}^-, w_{xz}^+, v, w_{v'x}^- | u)$$

$$= \max_{x,v,w's} P(w_{xy}^-, w_{xz}^+ | w_{v'x}^+ x, v, u) P(x, v, w_{v'x}^+ | u)$$

$$= \max_{x,v} [\lambda_y^*(x) \lambda_z^*(x) P(x|u,v) \max_{w_{v'x}^+} P(v, w_{v'x}^+ | u)]$$

Finally, using the marginal independence of $U$ and $W_{vx}^+$, we have

$$\lambda_x^*(u) = \max_{x,v} [\lambda_y^*(x) \lambda_z^*(x) P(x|u,v) \pi_x^*(v)] \quad (25)$$

In general, if $X$ has $n$ parents, $U_1, U_2, ..., U_n$, and $m$ children, $Y_1, Y_2, ..., Y_m$, node $X$ receives the messages $\pi_x^*(u_i), i=1,..., n$, from its parents and $\lambda_{y_j}^*(x), j=1,..., m$, from its children. Using the fixed probability $P(x|u_1, ..., u_n)$, processor $X$ will form the product

$$F(x, u_1, ..., u_n) =$$

$$\prod_{j=1}^m \lambda_{y_j}^*(x) P(x|u_1, ..., u_n) \prod_{i=1}^n \pi_x^*(u_i) \quad (26)$$

and then compute $n$ maximizations to obtain the messages destined for its parents:

$$\lambda_x^*(u_i) = \max_{x, u_k: k \neq i} [F(x, u_1, ..., u_n) / \pi_x^*(u_i)] \quad (27)$$

One additional maximization would be required to find the optimal value $x^*$ of $X$:

$$BEL^*(x) = \alpha \max_{u_k: 1 \leq k \leq n} F(x, u_1, ..., u_n) \quad (28)$$

$$x^* = \max_x^{-1} BEL(x) \quad (29)$$

Finally, the children-bound messages are computed by:

$$\pi_{y_j}^*(x) = \alpha \frac{BEL^*(x)}{\lambda_{y_j}^*(x)} \quad (30)$$

To complete the definition of the revision process, we must specify boundary conditions for nodes at the network periphery. These are determined from the probabilistic definition of the $\pi$ and $\lambda$ parameters and entail several cases:

1. *Anticipatory Node:* representing an uninstantiated variable with no successors. For such a node, $X$, we set $\lambda^*(x) = (1, 1, \cdots, 1)$, thus complying with Eqs. (13) and (14).

2. *Evidence Node:* representing a variable with instantiated value. Following Eq.(14), if variable $X$ assumes the value $x'$, we set

$$\lambda^*(x) = \begin{cases} 1 & \text{if } x = x' \\ 0 & \text{otherwise} \end{cases}$$

implying that, if $X$ has children, $Y_1,...,Y_m$, each child should receive the same message $\pi_{y_i}^*(x) = \lambda^*(x)$ from $X$.

3. *Dummy Node:* representing virtual or judgmental evidence. If node $X$ represents a virtual piece of evidence bearing on $U$, we post a $\lambda_x^*(u)$ message to $U$, where $\lambda_x^*(u) = K P(observation | u)$ and $K$ is any convenient constant.

4. *Root Node:* representing a variable with no parents. Eq.(15) dictates that, if $X$ has no parents, we set $\pi^*(x)$ to the prior probability of $X$, $P(x)$.

To prove that the propagation process terminates, we note that, since the network is singly-connected, every path must eventually end at either a root node having a single child or a leaf node having a single parent. Such single-port nodes act as absorption barriers; updating messages received through these ports get absorbed and do not cause subsequent updating of the outgoing messages. Thus, the effect of each new piece of evidence would subside in time proportional to the longest path in the network.

To prove that, at equilibrium, the selected values $w^*$ do, indeed, represent the most likely interpretation of the evidence at hand, we can reason by induction on the depth of the underlying tree, taking an arbitrary peripheral node as a root. The $\lambda^*$ or $\pi^*$ messages emanating from any leaf node of such a tree certainly comply with the definitions of Eqs. (14) and (15). Assuming that the $\lambda^*$ (or $\pi^*$) messages at any node of depth $k$ of the tree comply with their intended definitions of Eqs. (14 & 15), Eqs. (20-25) guarantee that they continue to comply at depth $k - 1$, and so on. Finally, at the root node, the overall $BEL^*(x^*)$ clearly represent $P(w^* | e)$, as in Eq.(11), which proves the assertion.

The propagation scheme described in this section bears many similarities to that used in belief updating (Eqs. (4-6)). In both cases, coherent global equilibria are obtained by local computations in time proportional to the network's diameter. Additionally, the messages $\pi^*$ and $\lambda^*$ bear both formal and semantic similarities to their $\pi$ and $\lambda$ counterparts, and the local computations required for updating them involve, roughly, the same order of complexity.

It is instructive, however, to highlight the major differences in the two schemes. First, belief updating involves *summation*, whereas in belief revision, *maximization* is the dominant operation. Second, belief updating involves more absorption centers than belief revision. In the former, every anticipatory node acts as an absorption barrier in the sense that it does not permit the passage of messages between its parents. This is clearly shown in Eq. (5); substituting $\lambda_y(x) = \lambda_z(x) = 1$ yields $\lambda_x(u) = 1$, which means that evidence in favor of one parent $(V)$ has no bearing on another parent $(U)$ as long as their common child $(X)$ receives no evidential support $(\lambda(x) = 1)$. This matches our intuition about how frames should interact; data about one frame (e.g., seismic data indicating the occurrence of an earthquake) should not evoke a change of belief about another unrelated frame (say, the possibility of a burglary in my home) just because the two may give rise to a common consequence sometimes in the future (e.g., triggering the alarm system). This frame-to-frame isolation no longer holds for belief revision, as can be seen from Eq. (25). Setting $\lambda_y^*(x) = \lambda_z^*(x) = 1$ still renders $\lambda_x^*(u)$ sensitive to $\pi_z^*(u)$.

Such endless frame-to-frame propagation raises both psychological and computational issues. Psychologically, in an attempt to explain a given phenomenon, the mere mental act of imagining the likely consequences of the hypotheses at hand will activate other, remotely related, hypotheses just because the latter could also cause the imagined consequence. We simply *do not encounter* that mode of behavior in ordinary reasoning; in trying to explain the cause of a car accident, we do not interject the possibility of lung cancer just because the two (accidents and lung cancer) could lead to the same eventual consequence -- death.

Computationally, it appears that, in large systems, the task of finding the most satisfactory explanation would require an excessive amount of computation; the propagation process would spread across loosely-coupled frames until every variable in the system reexamines its selected value $x^*$.

It turns out, though, that the propagation of the $\pi^* - \lambda^*$ parameters need not spread unchecked but can easily be confined to the relevant sections of the network by local suppressant mechanisms. If a satisfactory explanation is found in one section of the network, the $\lambda^* - \pi^*$ parameters emanating from that section would normally *reinforce* the existing $w^*$ values found outside that section and, whenever this occurs, further propagation will be arrested. For example, when we adopt the belief that a patient is suffering from cancer, we do also commit the belief that he/she will eventually be hospitalized. This commitment is often so strong that it is hard to distinguish between projected events and hard evidence in the sense that both would evoke a vivid picture of a hospital environment. However, this evocation would normally suppress, not activate, other possible causes of hospitalization, e.g., car accidents or food poisoning, because the $\lambda^*$ messages emanating from the hospital scene would reflect our full commitment to the original explanation -- cancer. Thus, once the processor responsible for handling the hospital scene realizes that the $\lambda^*$ message it is about to send toward the food-poisoning scene would only reinforce the currently held belief that no food poisoning took place, it will refrain from sending that message (until and unless contrary evidence develops), and this would keep the revision process from spreading beyond its natural boundaries.

### 4. Illustrating the Propagation Scheme

To illustrate the mechanics of the propagation scheme described in section 3, let us consider the diagnosis network of Fig. 3 [after Peng & Reggia, 1986], where the nodes at the top row, $\{d_1, d_2, d_3, d_4\}$, represent four hypothetical diseases and the nodes at the bottom row, $\{m_1, m_2, m_3, m_4\}$, four manifestations (or symptoms) of these diseases.

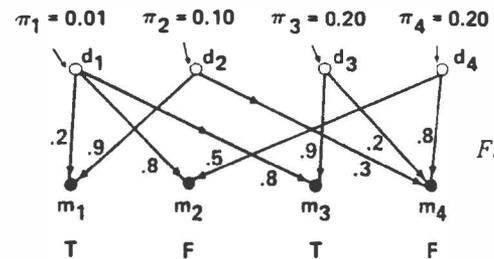

Figure 3

Let $d_i$ and $m_j$ denote the propositional variables associated with disease $d_i$ and manifestation $m_j$, respectively; each may assume a *TRUE* or *FALSE* value. Additionally, for each propositional variable $X$, we let $+x$ and $\neg x$ denote the propositions $X = TRUE$ and $X = FALSE$, respectively. Thus, for example,

$$P(\neg m_j \mid +d_i) = P(m_j = FALSE \mid d_i = TRUE)$$

would stand for the probability that a patient definitely having disease $d_i$ will *not* develop symptom $m_j$. The interaction among the possible causes of any given symptom is assumed to be of the "noisy OR-gate" type [Pearl 1985a], where the probability that a given symptom $m_j$ will be observed in the presence of disease set $D = \{d_i \mid i \in I_D\}$ is

$$P(+m_j \mid D) = P(+m_j \mid only\ diseases\ in\ D\ present)$$
$$= 1 - \prod_{i \in I_D} q_{ij} \qquad (31)$$

and $q_{ij}$ stands for

$$q_{ij} = P(\neg m_j \mid only\ d_i\ present) \qquad (32)$$

The two assumptions behind Eq.(31) are:

1. Symptom $m_j$ cannot occur in the absence of all diseases.

2. The mechanism which inhibits the occurrence of symptom $m_j$ in the presence of one disease is statistically independent of the mechanism which would inhibit it in the presence of another. Thus,

$$P(\neg m_j \mid D = \{d_1, d_2\}) =$$
$$P(\neg m_j \mid D = \{d_1\})\ P(\neg m_j \mid D = \{d_2\}) = q_{1j}\ q_{2j}$$

The link parameters $c_{ij} = 1 - q_{ij}$ are shown in the network of Fig. 3, together with the prior probabilities of the individual diseases, $\pi_i = P(+d_i)$.

A first step toward the analysis of such networks would be to rewrite the updating equations (26-30) for the specific OR-gate interaction given in Eq. (31) and (32). The function $F(x, u_1 \cdots u_n)$ obtains the form:

$$F(+x, u_1,...,u_n) = [1 - \prod_{i \in I_T} q_{ix}] \prod_{j=1}^{m} \lambda_{Y_j}^*(+x) \prod_{i=1}^{n} \pi_x^*(u_i) \quad (33)$$

$$F(\neg x, u_1,...,u_n) = \prod_{i \in I_T} q_{ix} \prod_{j=1}^{m} \lambda_{Y_j}^*(\neg x) \prod_{i=1}^{n} \pi_x^*(u_i) \quad (34)$$

where $I_T$ is the set of parents with value *TRUE*. The calculation of $BEL^*(x)$, Eq.(28), requires maximizing the two expressions of $F$ over all sets $I_T$. Maximizing the expression in Eq.(34) is readily done by noticing that adding element $k$ to $I_T$ merely amounts to multiplying the expression by a factor of $q_k \frac{\pi_{kx}}{1-\pi_{kx}}$, where $\pi_{kx}$ stands for

$$\pi_{kx} = \pi_x^*(+u_k) / [\pi_x^*(+u_k) + \pi_x^*(\neg u_k)] \quad (35)$$

Therefore, the optimal set of parents $I_T^*$ contains exactly those elements $k$ for which

$$q_{kx} \pi_{kx} > 1 - \pi_{kx} \quad (36)$$

For this optimized parent set we obtain

$F^*(\neg x, u \in I_T^*) =$

$$\prod_{j=1}^{m} \lambda_{Y_j}^*(\neg x) \prod_{i \in I_T^*} q_{ix} \pi_{ix} \prod_{i \notin I_T^*} (1-\pi_{ix}) \quad (37)$$

The expression in Eq.(33) presents a more difficult maximization problem because it is not in purely product form. The term $(1 - \prod_{i \in I_T} q_{ix})$ increases whenever we add another item to $I_T$, while the term

$$\prod_{i=1}^{n} \pi_x^*(u_i) = \prod_{i \in I_T} \pi_{ix} \prod_{i \notin I_T} (1-\pi_{ix}) \quad (38)$$

increases only when we add to $I_T$ element $k$ for which $\pi_{kx} > \frac{1}{2}$; otherwise, it decreases. Clearly, then, the optimizing set must contain every parent $u_k$ for which $\pi_{kx} > \frac{1}{2}$.

In general, the addition of element $k$ to $I_T$ would cause the expression in Eq.(33) to change by a factor of:

$$h_k = \frac{\pi_{kx}}{1-\pi_{kx}} \frac{(1-q_{kx} Q_I)}{1-Q_I} \quad (39)$$

where

$$Q_I = \prod_{i \in I_T} q_{ix} \quad (40)$$

Thus, $I_T$ can be improved by expansion if, and only if, there exist outside elements which make Eq.(39) greater than unity.

While there is no simple way of finding the optimal solution in the most general case, a reasonable, often optimal solution can be found with the following greedy algorithm:

1. Select all elements $i$ for which $\pi_{ix} \geq \frac{1}{2}$. (If none exist, select the element having the highest $1-q_{ix} \pi_{ix}/(1-\pi_{ix})$.) Call this set $I$, and compute $Q_I$ as in Eq.(40).

2. For each of the remaining elements, $k$, calculate a merit function $h_k$, as in Eq.(39);

3. if all $h_k \leq 1$, return $I$ and quit; else,

4. add to $I$ the element with the highest $h_k$, and delete all elements with $h_k \leq 1$;

5. recompute $Q_I$, and repeat from step 2.

Peng and Reggia [1986] developed an admissible heuristic search algorithm for finding the exact optimal set $I_T$ in a more general setting, optimizing over the parents of all observed nodes. However, since our task focusses on relatively small groups of parents, each group sharing a single common child, such powerful techniques will not be necessary.

Let us return to the diagnosis network of Fig. 3 and imagine that the patient at hand is definitely suffering from disease $d_1$, showing symptoms $\{m_1, m_3\}$ but *not* of $\{m_2, m_4\}$. Our task is to find the disease *combination* most likely to explain the observed data, namely, to find a subset of $\{d_2, d_3, d_4\}$ which, together with $+d_1$ and the evidence $e = \{+m_1, \neg m_2, +m_3, \neg m_4\}$, constitutes the most probable instantiation of all variables.

For convenience, let us adopt the following notation:

$$\lambda_{ji} = \lambda_{m_j}^*(+d_i) / \lambda_{m_j}^*(\neg d_i) \quad (41)$$

$$\pi_{ij} = \pi_{m_j}^*(+d_i) / [\pi_{m_j}^*(+d_i) + (\pi_{m_j}^*(\neg d_i)] \quad (42)$$

Thus,

$$\lambda_{m_j}^*(d_i) = [(\lambda_{m_j}^*(+d_i), \lambda_{m_j}^*(\neg d_i))] = \alpha'(\lambda_{ji}, 1) \quad (43)$$

$$\pi_{m_j}^*(d_i) = [(\pi_{m_j}^*(+d_i), \pi_{m_j}^*(\neg d_i))] = \alpha''(\pi_{ij}, 1-\pi_{ij}) \quad (44)$$

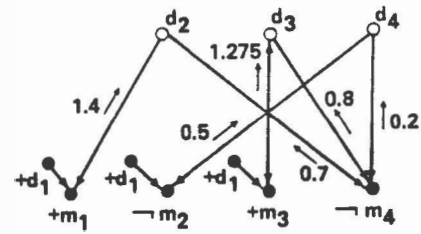

Figure 4(a)

The network in Fig. 3 becomes singly-connected upon instantiating $d_1$. Fig. 4(a) shows its message-passing topology, together with the initial messages posted by the instantiated variables:

$\lambda_{12} = (1-q_{11} q_{21}) / (1-q_{11}) = (1-0.8 \cdot 0.9) / (1-0.8) = 1.400$
$\lambda_{33} = (1-q_{13} q_{33}) / (1-q_{13}) = (1-0.2 \cdot 0.1) / (1-0.2) = 1.225$
$\lambda_{24} = q_{12} q_{42} / q_{12} = 0.5$
$\lambda_{44} = q_{44} = 0.2$
$\lambda_{43} = q_{34} = 0.8$
$\lambda_{42} = q_{24} = 0.7$

The calculation of $\lambda_{4i}$ is based on the fact that, for all $i$, $\pi_{i4} / (1 - \pi_{i4})$ is smaller than unity; so, Eq.(27) is maximized by setting all $u_k$ to *FALSE*.

206

At the second phase, each $d_i$ processor inspects the messages posted on its links and performs the operations specified in Eq.(22) or Eq.(30). This leads to the network in Fig. 4(b), with:

$\pi_{24} = \pi_2 \lambda_{12} / (1 - \pi_2 + \pi_2 \lambda_{12}) = 0.135$
$\pi_{34} = \pi_3 \lambda_{33} / (1 - \pi_3 + \pi_3 \lambda_{33}) = 0.234$
$\pi_{44} = \pi_4 \lambda_{24} / (1 - \pi_4 + \pi_4 \lambda_{24}) = 0.111$
$\pi_{21} = \pi_2 \lambda_{42} / (1 - \pi_2 + \pi_2 \lambda_{42}) = 0.072$
$\pi_{33} = \pi_3 \lambda_{43} / (1 - \pi_3 + \pi_3 \lambda_{43}) = 0.167$
$\pi_{42} = \pi_4 \lambda_{44} / (1 - \pi_4 + \pi_4 \lambda_{44}) = 0.048$

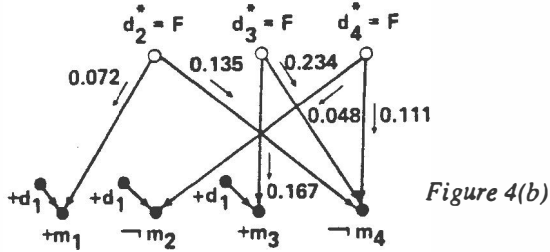

Figure 4(b)

The $x^*$ value chosen by each of the $d_i$ processors is FALSE, (See Eq. (28)) because, for each $i = 2, 3, 4$, we have

$$\frac{BEL^*(+d_i)}{BEL^*(\neg d_i)} = \prod_{j=1}^{4} \lambda_{ji} \frac{\pi_i}{1 - \pi_i} < \tfrac{1}{2}$$

For example, processor $d_2$ receives: $\lambda_{12} = 1.4, \lambda_{42} = 0.7$; so,

$$\frac{BEL^*(+d_2)}{BEL^*(\neg d_2)} = \frac{\lambda_{12} \cdot \lambda_{42} \cdot \pi_2}{1 \cdot 1 \cdot (1-\pi_2)} = \frac{1.4 \cdot 0.7 \cdot 0.2}{1 \cdot 1 \cdot 0.8} = 0.245 < \tfrac{1}{2}.$$

The messages $\pi_{21}, \pi_{33}$ and $\pi_{42}$ will eventually get absorbed at node $d_1$, while $\pi_{24}, \pi_{34}$ and $\pi_{44}$ are now posted on the ports entering node $m_4$. Again, none of these messages meet the criterion $\pi_{i4} q_{i4} > 1 - \pi_{i4}$, which would qualify its sender to be set TRUE in the maximization of Eq.(27). Consequently, the $\lambda^*$ messages generated by node $m_4$ on the next activation phase remain unchanged (Fig. 4(c)), and the process halts with the current $w^*$ values: $d_2 = d_3 = d_4 = FALSE$.

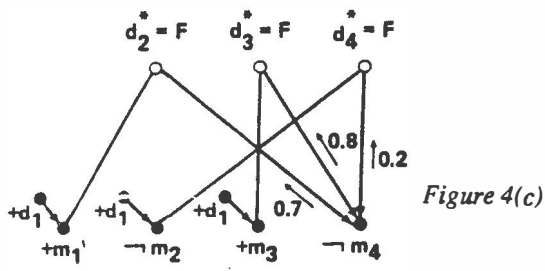

Figure 4(c)

To demonstrate the spread of belief revision, let us imagine that we now retract the assumption $d_1 = TRUE$; instead (perhaps after re-examining the data), we posit the converse: $d_1 = FALSE$. This change of opinion results in the messages $\pi_{11} = \pi_{12} = \pi_{13} = 0$ being posted on all those links emanating from node $d_1$ which get translated to $\lambda_{12} = \infty, \lambda_{13} = \infty$, and $\lambda_{24} = q_{21} = .5$. This means that $d_2$ and $d_3$ will switch simultaneously and permanently to state TRUE while $d_4$, by virtue of

$$\frac{BEL^*(+d_4)}{BEL^*(\neg d_4)} = \lambda_{24} \cdot \lambda_{44} \cdot \pi_4 / (1 - \pi_4)$$

$= 0.50 \cdot 0.20 \cdot 0.20 / 0.80 = .025 < \tfrac{1}{2},$

tentatively remains at the state FALSE, as illustrated in Fig. 5(a).

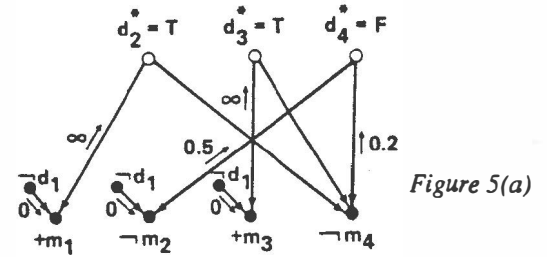

Figure 5(a)

During the next activation phase (Fig. 5(b)), $d_2$ and $d_3$ post the messages $\pi_{24} = \pi_{34} = 1$, which $m_4$ inspects for possible updating of $\lambda_{44}$. However, these new messages will not cause any change in $\lambda_{44}$ because, according to Eqs. (27) and (31), the ratio $\lambda_{44}$ remains

$$\lambda_{44} = \frac{P(\neg m_4 \mid +d_4, d_2, d_3)}{P(\neg m_4, \mid d_2, d_3)} = q_{44},$$

independent of the states of $d_2$ and $d_3$.

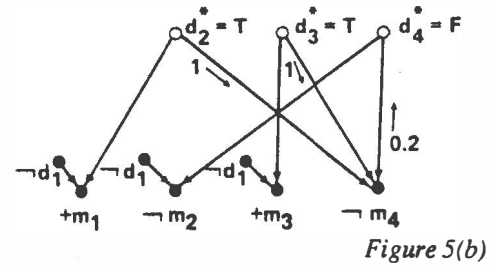

Figure 5(b)

Thus, under the current premise $\neg d_1$, the most likely interpretation of the symptoms observed is $\{+d_2, +d_3, \neg d_4\}$. In view of the network topology and the evidence observed, we intuitively expect this conclusion. Additionally, the $\lambda^*$ and $\pi^*$ messages calculated can be used to mechanically generate a verbal explanation such as the following:

> "Since we have ruled out disease $d_1$, the only possible explanation for observing symptoms $m_1$ and $m_3$ is that the patient suffers, simultaneously, from $d_2$ and $d_3$. The fact that $m_2$ and $m_4$ both came out negative indicates that disease $d_4$ is absent."

## 5. Coping with Loops

Loops are undirected cycles in the underlying network, i.e., the Bayesian network without the arrows. When loops are present, the network is no longer singly-connected, and local propagation schemes invariably run into trouble. The two major methods for handling loops while still retaining some of the flavor of local computation are: *clustering* and *conditioning*.

Clustering involves the forming of compound variables in such a way that the topology of the resulting network is singly-connected. For example, if in the network of Fig. 1 we define the compound variables:

$$Y_1 = \{X_1, X_2\} \quad Y_2 = \{X_2, X_3\}$$

207

the following tree ensues: $X_4 \leftarrow Y_1 \rightarrow Y_2 \rightarrow X_5 \rightarrow X_6$.

In the network of Figure 3, defining the variables $D_{234} = \{d_2, d_3, d_4\}$ and $M_{123} = \{m_1, m_2, m_3\}$, we obtain a singly-connected network of the form:

$$d_1 \rightarrow M_{133} \leftarrow D_{234} \rightarrow m_4$$

Alternatively, the clusters could be made to overlap each other until they entail all the links of the original network, and the interdependencies between any two clusters are mediated solely by the variables which they share. For example, in the network of Fig. 1, if we define $Z_1 = \{X_1, X_2, X_4\}$, $Z_2 = \{X_1, X_2, X_3\}$, $Z_3 = \{X_2, X_3, X_5\}$ and $Z_4 = \{X_5, X_6\}$, the dependencies among the $Z$ variables will be described by the chain

$$Z_1 \text{---} \{x_1, x_2\} \text{---} Z_2 \text{---} \{x_2, x_3\} \text{---} Z_3 \text{---} \{x_5\} \text{---} Z_4$$

where the $x$ symbols on the links identify the set of $X$ variables common to any pair of adjacent $Z$ clusters [see Lemmer 1983; Spiegelhalter 1985].

These clustered networks can be easily processed with the propagation techniques of Section 3, except that the multiplicity of each compound variable increases exponentially with the number of elementary variables it contains. Consequently, the size of either the link matrices or the messages transmitted may become prohibitively large.

An extreme case of clustering would be to represent all ancestors of the observed findings by *one* compound variable. Assigning a definite value to this compound variable would constitute an explanation for the findings observed. Indeed, this is the approach taken by Cooper [1984] and Peng & Reggia [1986]. To search for the best explanation through the vast domain of possible values associated with the explanation variable, admissible heuristic strategies had to be devised, similar to that of the $A^*$ algorithm [Pearl 1984]. The main disadvantage of this technique is the loss of conceptual flavor; the optimization procedure does not reflect familiar mental processes and, consequently, it is hard to construct meaningful arguments to defend the final conclusions.

Conditioning is an attempt to preserve, as much as possible, the conceptual nature of the interpretation process, by performing the major portion of the optimization using local computations *at the knowledge level itself*, i.e., using the links provided by the network as communication channels between simple, identical, autonomous and semantically related processors.

The basic idea behind conditioning was already illustrated in the example of section 4 where the instantiation of variable $d_1$ rendered the rest of the network singly-connected, amiable to the propagation technique of section 3. We saw that the assumption $+d_1$ yields the interpretation $\{\neg d_2, \neg d_3, \neg d_4\}$, while $\neg d_1$ yields $\{+d_2, +d_3, \neg d_4\}$. The question now is to decide which of the two interpretations is more plausible or, in other words, which has the highest posterior probability given the evidence $e = \{+m_1, \neg m_2, +m_3, \neg m_4\}$ at hand. A direct way to decide between the two candidates is to calculate the two posterior probabilities, $P(I^+|e)$ and $P(I^-|e)$, where

$$I^+ = \{+d_1, \neg d_2, \neg d_3, \neg d_4\} \text{ and } I^- = \{\neg d_1, +d_2, +d_3, \neg d_4\}.$$

These calculations are quite simple, because instantiating the $d$ variables *separates* the $m$ variables from each other, so that the posterior probabilities involve only products of $P(m_j | parents \ of \ m_j)$ over the individual symptoms and a product of the prior probabilities over the individual diseases. For example,

$$\begin{aligned}P(I^+|e) &= \alpha P(I^+) P(e|I^+) \\ &= \alpha \pi_1(1-\pi_2)(1-\pi_3)(1-\pi_4)(1-q_{11}) q_{12}(1-q_{13}) \\ &= \alpha \cdot 0.01 \cdot 0.90 \cdot 0.80 \cdot 0.80 \cdot 0.20 \cdot 0.90 \cdot 0.80 \\ &= \alpha \, 8.2944 \times 10^{-4}\end{aligned}$$

$$\begin{aligned}P(I^-|e) &= \alpha P(I^-) P(e|I^-) \\ &= \alpha (1-\pi_1) \pi_2 \pi_3(1-\pi_4)(1-q_{21})(1-q_{33}) q_{24} q_{34} \\ &= \alpha \cdot 0.99 \cdot 0.10 \cdot 0.20 \cdot 0.80 \cdot 0.90 \cdot 0.90 \cdot 0.70 \cdot 0.80 \\ &= \alpha \, 7.18 \times 10^{-3}\end{aligned}$$

Since $\alpha = [P(e)]^{-1}$ is a constant, we conclude that $I^-$ is the most plausible interpretation of the evidence $e$.

These two globally-supervised computations are identical to those used by Peng & Reggia [1986]. However, we use them to compare only two candidates from the space of $2^4$ possible disease combinations. Most of the interpretation work was conducted by local propagation, selecting the appropriate match for each of the two assumptions $+d_1$ and $\neg d_1$. Thus, we see that, even in multiply-connected networks, local propagation provides computationally effective and conceptually meaningful method of trimming the space of interpretations down to a manageable size.

The effectiveness of conditioning depends heavily on the topological properties of the network. In general, a set of several nodes (called a cycle cutset) must be instantiated before the network becomes singly-connected. This means that $2^c$ candidate interpretations will be generated by local propagation, where $c$ is the size of the cycle cutset chosen for conditioning. Since each propagation phase takes only time linear with the number of variables in the system $(n)$, the overall complexity of the optimal interpretation problem is exponential with the size of the cycle cutset that we can identify. If the network is sparse, topological considerations can be used to find a small cycle-cutset and render the interpretation task tractable. Although the problem of finding the minimal cycle cutset is NP hard, simple heuristics exist for finding close-to-minimal sets [Levy & Low, 1983]. Identical complexity considerations apply to the task of belief updating [Pearl 1985b], which show that finding the globally best explanation is no more complex than finding the degree of belief for any individual proposition.

It is interesting to note that there is a definite threshold value for $\pi_1$, $\pi_1 = 0.0804$, at which the two interpretations $I^+$ and $I^-$ have equal likelihood. That means that, as evidence in favor of $+d_1$ accumulates and $\pi_1$ increases beyond the value 0.0804, the system will switch abruptly from interpretation $I^-$ to interpretation $I^+$. This abrupt "change of view" is a collective phenomenon, characteristic of massively parallel systems, and is reminiscent of the way people's beliefs undergo complete reversal in response to a minor clue. Note, though, that the transition is reversible, i.e., as $\pi_1$ decreases, the system will switch back to the interpretation $I^-$ at exactly the same threshold value, $\pi_1 = 0.0804$. No hysteresis occurs because, although the computations are done locally, $w^*$ is globally optimal and is, therefore, a unique function of all systems' parameters.

This reversibility differs from human behavior in that, once we commit our belief to a particular interpretation, it often takes more convincing evidence to make us change our mind than the evidence which got us there in the first place. Simply discrediting a piece of evidence would not, in itself, make us abandon the beliefs which that evidence induced [Ross & Anderson, 1982; Harman, 1986]. The phenomena is very pronounced in perceptual tasks; once we adopt one view of Necker's cube or an Escher sketch, it takes a real effort to break ourselves loose and adopt alternative interpretations. Irreversibility (or hysteresis) of that kind is characteristic of systems with local feedback. For example, if the magnetic spin of one atom heads north, it sets up a magnetic field which encourages its neighbors to follow suit; when the neighbors' spins eventually turn north, they generate a magnetic field which further "locks" the original atom in its north-pointing orientation.

The source of hysteresis in human belief revision is not yet clear. One possibility is that local feedback loops are triggered between evoked neighboring concepts; e.g., if I suspect fire, I expect smoke, and that very expectation of smoke reinforces my suspicion of fire -- as if I actually saw smoke. This is a rather unlikely possibility because it would mean that even in simple cases (e.g., the fire and smoke example), people are likely to confuse internal thinking with genuine evidence. A more reasonable explanation is that, by-and-large, the message-passing process used by people is feedback-free and resembles that of Section 3, where the $\pi^*$ and $\lambda^*$ on the same link are orthogonal to each other. However, in complex situations, where loops are ramparts, people simply cannot afford the overhead computations required by conditioning or clustering. As an approximation, then, they delegate the optimization task to local processes and continue to pass messages as if the belief network were singly-connected. The resultant interpretation, under these conditions, is locally, not globally, optimal, and this accounts for the irreversibility of belief revision.

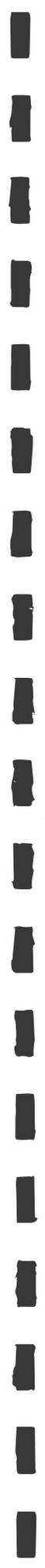